\documentclass[letterpaper, 10 pt, conference]{ieeeconf}  % Comment this line out if you need a4paper

\IEEEoverridecommandlockouts                              % This command is only needed if 
\usepackage{graphics} % for pdf, bitmapped graphics files
\usepackage[table,xcdraw]{xcolor}
\usepackage{epsfig} % for postscript graphics files
\usepackage{mathptmx} % assumes new font selection scheme installed
\usepackage{times} % assumes new font selection scheme installed
\usepackage{amsmath} % assumes amsmath package installed
\usepackage{amssymb}  % assumes amsmath package installed
\usepackage{floatrow}
\usepackage{subcaption}

\title{\LARGE \bf
Monocular Visual Odometry for an Unmanned Sea-Surface Vehicle 
}

\author{George Terzakis$^{1}$, Riccardo Polvara$^{2}$, Sanjay Sharma$^{2}$, Phil Culverhouse$^{2}$ and Robert Sutton$^{2}$% <-this % stops a space
\thanks{$^{1}$George Terzakis is with the University of Portsmouth, PO1 3HF, UK. Email: {\tt\small george.terzakis@port.ac.uk}}%
\thanks{$^{2}$Riccardo Polvara, Sanjay Sharma, Phil Culverhouse and Robert Sutton are with Plymouth University, PL4 8AA, UK. Emails: {\tt\small \{riccardo.polvara, sanjay.sharma, p.culverhouse, r.sutton\}@plymouth.ac.uk}}
}

\begin{document}

\maketitle
\thispagestyle{empty}
\pagestyle{empty}

%%%%%%%%%%%%%%%%%%%%%%%%%%%%%%%%%%%%%%%%%%%%%%%%%%%%%%%%%%%%%%%%%%%%%%%%%%%%%%%%
\begin{abstract}

We tackle the problem of localizing an autonomous sea-surface vehicle in river estuarine areas using monocular camera and angular velocity input from an inertial sensor. Our method is challenged by two prominent drawbacks associated with the environment, which are typically not present in standard visual simultaneous localization and mapping (SLAM) applications on land (or air): a) Scene depth varies significantly (from a few meters to several kilometers) and, b) In conjunction to the latter, there exists no ground plane to provide features with enough disparity based on which to reliably detect motion. To that end, we use the IMU orientation feedback in order to re-cast the problem of visual localization without the mapping component, although the map can be implicitly obtained from the camera pose estimates. We find that our method produces reliable odometry estimates for trajectories several hundred meters long in the water. To compare the visual odometry estimates with GPS based ground truth, we interpolate the trajectory with splines on a common parameter and obtain position error in meters recovering an optimal affine transformation between the two splines. 

\end{abstract}

%%%%%%%%%%%%%%%%%%%%%%%%%%%%%%%%%%%%%%%%%%%%%%%%%%%%%%%%%%%%%%%%%%%%%%%%%%%%%%%%
\section{INTRODUCTION}

Outdoor visual SLAM is a problem under constant scrutiny in the robotics research community \cite{dunkley2014visual, schleicher2009real, konolige2010large, faessler2016autonomous}. However, to the best of our knowledge, there have been not many cases in which the SLAM algorithm targets sequences of estuarine and possibly, natural scenes from the vantage point of a surface vehicle. Perhaps the only known recorded case is the work of \cite{griffith2015robot} on lakeshore monitoring which makes use of visual SLAM to localize the vehicle on the surface of the lake. 
The approach of Griffith et al. relies on the same tools as the method described in this paper. The idea is to detect a sparse set of Harris corners \cite{harris1988combined} in an image and thereafter compute camera pose by tracking these features using the pyramidal Lucas – Kanade tracker \cite{lucas1981iterative, bouguet2001pyramidal}. Of course, the primary objective of Griffith’s work is the registration of shore images and vehicle localization is simply a subsidiary operation used to confine the search for an image within a subset of images taken in a region close to the estimated position of the boat. It follows that multiple poses can be refined without time limitations in large-scale bundle adjustment runs. 
\par The work in this paper is motivated by the scenario of a surface vehicle \cite{naeem2006design} cruising autonomously on GPS feedback which may be interrupted; therefore visual odometry on lakeshore images is used as an auxiliary localization system for the vehicle during periods in which GPS reception is disabled. Thus, in contrast with Griffith et al., this is an exclusively real-time visual SLAM problem and the exact same approach cannot be employed due to the inherent limitations in computational resources. There are however time-efficient alternatives which employ non-linear optimization over only a small pool of poses and images in a sliding time-window, such as local bundle adjustment \cite{klein2009parallel} and generally provide satisfactory pose estimates on demand. 
\par
\begin{figure}[ht]
\centering
\centerline{\includegraphics[width=\textwidth,keepaspectratio]{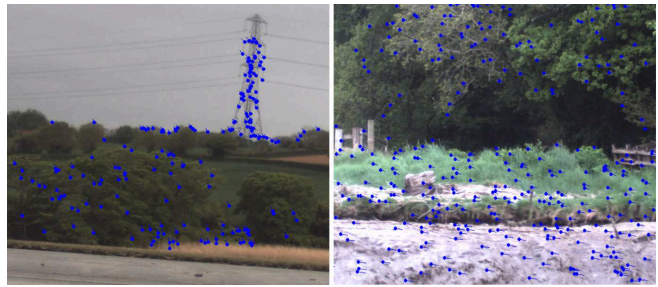}}
\caption{Extreme opposites in estuarine sequences. On the left, only very distant features over the horizon can be tracked, while on the right, scene depth is very small. In the first case, disparity is very small and only a few carefully selected correspondences can determine motion; in the second case, the disparity is sufficient to determine motion from the majority of features.}
\label{fig:sample_shore_images}       % Give a unique label
\end{figure}

 One other significant limitation in our application has to do with extreme depth variations in the shore sequence (\ref{fig:sample_shore_images}). Although there are ways to assess the nature of the photogrammetric degeneracies in the geometry of two views \cite{torr1995robust,kanatani1998geometric,torr1998robust, hartley2000ambiguous}, these methods require Monte-Carlo based model selection and would impose a significant overhead if applied on a regular basis. One remedy would be to include ground-plane features in the field of view \cite{konolige2010large, leutenegger2015keyframe, lourakis2015planetary} and allow motion estimation to be affected primarily by these features. Unfortunately, this is generally not an option when the camera is on-board a surface vehicle due to the presence of the water-surface instead of the ground. It therefore becomes necessary to infer camera motion solely based on the features in the background beyond the sea-surface.
\par
\begin{figure}[ht]
\centering
\centerline{\includegraphics[width=\textwidth,keepaspectratio]{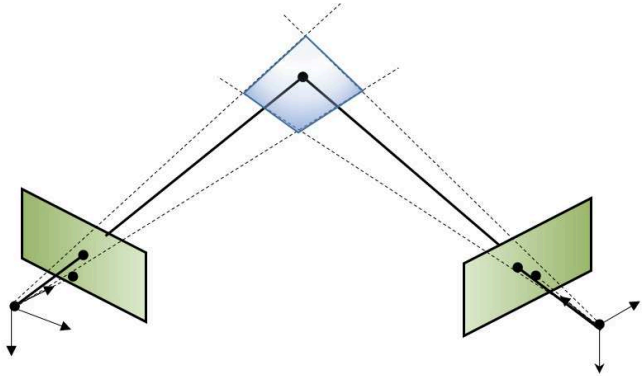}}
\caption{Uncertainty induction with depth. The size of the blue shaded rhombic region surrounding the triangulated map point is representative of its uncertainty. A similar sketch can be found in Multiple View Geometry in Computer Vision (Hartley and Zisserman 2003).}
\label{fig:noisy_triangulation}       % Give a unique label
\end{figure}

 Another significant problem that arises from extreme depth variation is the fact that the map can be potentially harmful for pose prediction, due to tracking noise induction with large depth. The problem of reconstructing points with very small disparity is that tracking noise is augmented proportionally to depth as shown in Figure \ref{fig:noisy_triangulation}. And since having exclusively distant points in the map is a very likely case as illustrated on the left of Figure \ref{fig:sample_shore_images}, it will be difficult to assess the quality of the position estimates in order to choose inliers for pose estimation with a Perspective-n-Pose (PnP) algorithm. In short, the map points obtained from scenes with large scene depth can jeopardize SLAM if used for pose prediction.

\section{Problem Setup}

\subsection{Overview}

\begin{figure}[ht]
\centering
\centerline{\includegraphics[width=\textwidth,keepaspectratio]{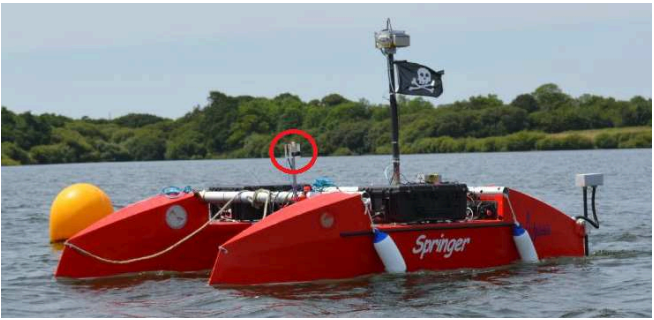}}
\caption{The unmanned surface vehicle. The camera (circled in red) is mounted on the left hull, pointing sideways.}
\label{fig:springer}       % Give a unique label
\end{figure}

The method proposed herein aims at estimating the pose of a sea-surface vehicle in the context of GPS-denied scenario using an on-board camera with view of the shore. In particular, in the case of GPS signal loss, the vehicle should be able to continue cruising autonomously based solely on visual feedback until GPS becomes available again. We focus on river estuarine environments where the primary assumption of shore visibility is met. The camera is mounted on the side of the vehicle as shown in Figure \ref{fig:springer}.

\subsection{Apparatus}

All video sequences were obtained using a Pointgrey Flea3 USB3 camera, typically at a resolution of 800×600 pixels. We addressed camera calibration with Zhang’s chessboard method \cite{zhang1999flexible,zhang2000flexible} as implemented in OpenCV computer vision library \cite{bradski2000opencv}. For the rest of this paper, it is assumed that the camera is calibrated and all formulas involving projections are given in normalized Euclidean coordinates. 
\begin{figure}[ht]
\centering
\centerline{\includegraphics[width=\textwidth,keepaspectratio]{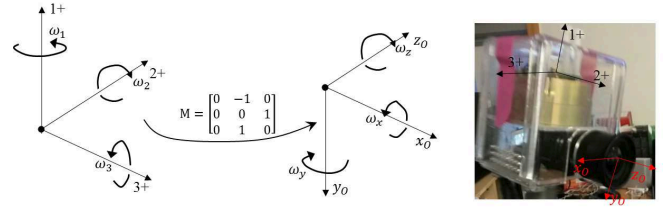}}
\caption{Flea3 and SiIMU02 (right). The SiIMU02 coordinate frame (left) is shown with the 3 axes labelled 1+, 2+ and 3+ and M denotes the transformation which transforms the sampled angular velocity vectors with the camera frame (shown in red).}
\label{fig:SiIMU02}       % Give a unique label
\end{figure}
\par Inertial data were sampled using a Goodrich SiIMU02  (Figure \ref{fig:SiIMU02}) at a rate of 150 Hz. It should be noted that the SiIMU02 coordinate frame does not represent a right handed coordinate system and therefore it is necessary to apply a transformation M to the samples in order to obtain the angular velocities $ω_x$, $ω_y$, $ω_z$ about the axes $x$, $y$, $z$ of the local camera frame. Thus, the rotation matrix representing the change in IMU orientation is:

\begin{equation}
\label{eq:rodrigues}
R = I_3 + \frac{\sin{\Vert\omega\Vert}}{\Vert\omega\Vert}\left[\omega\right]_{\times} + \frac{1 - \cos{\Vert\omega\Vert}}{{\Vert\omega\Vert}^ 2} {\left[\omega\right]}^2_{\times}
\end{equation}
where $\omega={\begin{bmatrix} \omega_x & \omega_y & \omega_z \end{bmatrix}}^T$ is the sampled angular velocity vector. We addressed the problem of minor misalignments between the camera the IMU using the recovered camera calibration extrinsic parameters as described in camera-aided robot calibration \cite{zhuang1996camera}.

\section{Method}
\begin{figure}[ht]
\centering
\centerline{\includegraphics[width=\textwidth,height=0.45\textwidth]{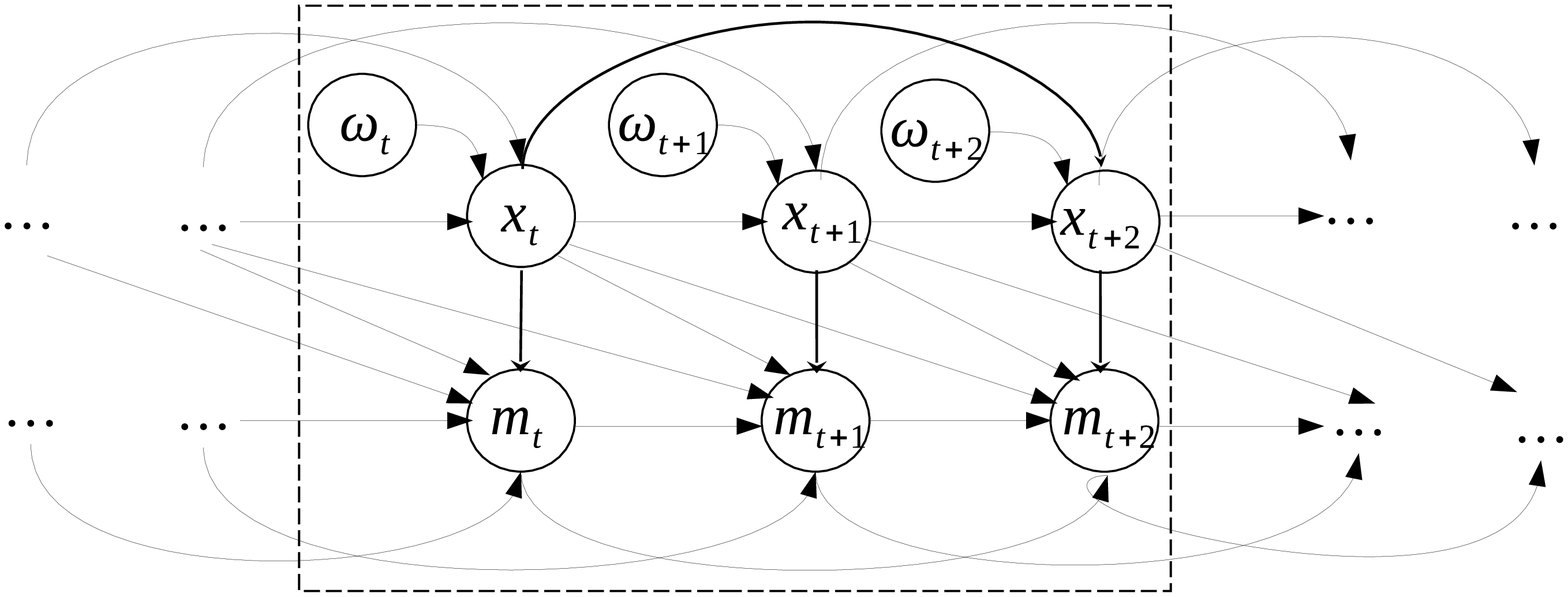}}
\caption{A Bayes network depiction of the SLAM paradigm adaptation of our method. The dashed-line rectangle indicates a 3-frame sliding window from time $t$ to time $t+2$.}
\label{fig:SLAM_BayesNet}       % Give a unique label
\end{figure}

Our method loosely follows the standard SLAM pipeline of pose prediction and refinement with measurements over a sliding window of no more than 5 frames. The idea is to detect new features in each frame and track them for up to 5 frames in the sequence. However, pose is estimated directly on the pairs of corresponding Euclidean projections instead of using the reconstructed map-points. In effect, this is a version of SLAM in which the map can only be inferred from camera pose in conjunction with the measurements and is not directly involved in pose prediction. Figure \ref{fig:SLAM_BayesNet} illustrates our adaptation of the SLAM paradigm in the form of a Bayes network. The pose at time instance-$t$ (time is discrete) is a vector $x_t = {\begin{bmatrix} s_t & \psi_t \end{bmatrix}}^T$ containing the position of the camera $s_t$ and its orientation parameter vector $\psi_t$. We also denote the IMU angular velocity sample with $\omega_t$ and the normalized Euclidean projections of the tracked features with $m_t$.

\par In the SLAM adaptation of Figure \ref{fig:SLAM_BayesNet}, each tracked feature location at time $t$ is regarded as a function of its projected location and camera pose in the view in which it was originally detected; and since we are using a sliding window of length 3, it follows that it can only be associated with times $t-1$ or $t-2$, hence the factor between $m_t$, $x_{t-1}$, $x_{t-2}$, $m_{t-1}$, $m_{t-2}$ in the network. The measurement model is simply the pinhole projective relationship between the i\textsuperscript{th} feature measurement $m^{\left(i\right)}_t$ at time $t$ and its corresponding image in the view of original detection (henceforth called the “home view”) at time $h$:

\begin{equation}
\label{eq:general_projection_model}
m^{\left(i\right)}_t = \frac{R_{h_i} R{^T}_t \left( Z_{h_i} m^{\left(i\right)}_{h_i} - R_h\left(s_t -s_{h_i}\right)\right) }{1^T_z R_{h_i} R^T_t \left(Z_{h_i} m^{\left(i\right)}_{h_i} - R_{h_i} \left(s_t - S_{h_i}\right)\right)}
\end{equation}
where $R_t$ is the rotation matrix corresponding to the camera frame orientation at time $t$ (direction vectors are stored column-wise), $s_t$ is the position of the camera at time t in world coordinates, $h_i$ is the time index of the home view (i.e., original detection) if the i\textsuperscript{th} feature, $Z_t$ is the depth of the map-point at time $t$ and $1_z={\begin{bmatrix} 0 & 0 & 1 \end{bmatrix}}^T$. As will be shown, provided that orientation is known, it is possible to eliminate $Z_h$ from eq. \ref{eq:general_projection_model} and replace it with an expression that contains $s_t$, $s_h$, $R_t$, $R_h$ and the measurements. Thus, the corresponding conditional distribution of $m_t$ depends only on previous measurements and poses and it can be loosely regarded as the marginal of the standard visual SLAM measurement model over the map.

\subsection{Predicting camera pose} 
Our method makes use of a technique that loosely draws inspiration from the work of Kneip et al. \cite{kneip2011robust}. In particular, knowing the rotation matrix between two key-frames, we are able to do camera resectioning directly from image correspondences, thereby circumventing the potentially noisy distant map-points. To lighten notation in the derivations that follow, we isolate two views from the sequence in both of which a feature is measured and assume, without constraining generality, that the first camera is at the origin and its orientation is the identity. The relationship of eq. \ref{eq:relative_projection_model} can now be re-written as follows:
\begin{equation}
\label{eq:relative_projection_model}
m_2 = \frac{R^T (Z_1 m_1 -b)}{1^{T}_Z R^T (Z_1 m_1 -b)}
\end{equation}

where $m_2$ and $m_1$ are the measured normalized Euclidean coordinates of the feature in the current and previous (home) view, $Z_1$ is the depth of the feature in the previous (home) view, $b$ is the baseline vector in the coordinate frame of the previous camera and $R$ is the relative orientation matrix (containing the current camera frame in column-wise arrangement).
\begin{figure}[ht]
\centering
\centerline{\includegraphics[width=\textwidth,keepaspectratio]{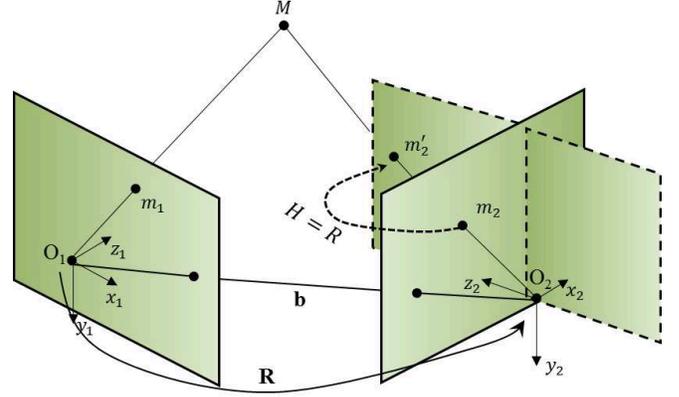}}
\caption{Removing rotational effect from the projections in the second view. The rotation matrix $R$ transforms the first camera frame triad $\left( x_1, y_1,z_1\right)$ at $O_1$ into the second, $\left(x_2, y_2, z_2\right)$ attached to the camera center $O_2$. The homography $H=R$ is applied to the normalized Euclidean projection $m_2$ in order to obtain its "unrotated" version, $m^{\prime}_2$ in the virtual view (shown with dashed outline).}
\label{fig:feature_unprojection}       % Give a unique label
\end{figure}

\par Now, provided reliable prior information on relative orientation between two camera frames, it is possible to remove the effects of rotational motion from the image projections in the current view by creating a new, virtual view in which all projections are, the result of pure translational motion. This way, motion equations become linear in the translation components and standard least squares optimization can be applied. Figure \ref{fig:feature_unprojection} illustrates this concept of “unrotating” projections in order to produce a virtual view that shares the same baseline with the original, but without the rotational portion of projective distortion. Specifically, if a pure rotation $R^T$ is applied to the second camera frame, then it will align with the first. It follows that the normalized projections in the “unrotated” view will simply transform by a homography $H=R$:

\begin{equation}
\label{eq:rotated_points}
m^{\prime}_2 = \frac{Rm_2}{1^T_z R m_2}
\end{equation}
where $\propto$ denotes equality up-to-scale. Eq. \ref{eq:relative_projection_model} can now be re-written without the rotation matrix using the “unrotated” projection $m^{\prime}_2$:

\begin{equation}\label{eq:unrotated_projection_model}
m^{\prime}_2 = \frac{Z_1 m_1 - b}{1^T_z ( Z_1 m_1 - b)}
\end{equation}
Equation \ref{eq:unrotated_projection_model} decomposes in the following two equations corresponding to the projections in $x$ and $y$ axes:
\begin{eqnarray}
\label{eq:LS_equations_equation_1}
(x^{\prime}_2 - x_1)Z_1 &=& x^{\prime}_2 b_z - b_x \\
\label{eq:LS_equations_equation_2}
(y^{\prime}_2 - y_1)Z_1 &=& y^{\prime}_2 b_z - b_y
\end{eqnarray}
where $m^{\prime}_2 = {\begin{bmatrix}x^{\prime}_2 & y^{\prime}_2 & 1\end{bmatrix}}^T$, $m_1 = {\begin{bmatrix}x_1&  y_1&  1\end{bmatrix}}^T$ and $b = {\begin{bmatrix} b_x&  b_y&  b_z\end{bmatrix}}^T$. The depth can be eliminated if we simply multiply eq. \ref{eq:LS_equations_equation_1} by $\left(y^{\prime}_2 - y_1\right)$ and eq. \ref{eq:LS_equations_equation_2} by $\left(x^{\prime}_2 - x_1\right)$  and then, subtract eq. \ref{eq:LS_equations_equation_2} from eq. \ref{eq:LS_equations_equation_1}:
\begin{equation}
\begin{split}
\label{eq:LS_equation_without_depth}
-\left(y^{\prime}_2 -y_1\right)b_x +& \left(x^{\prime}_2 - x_1\right)b_y + \\ & \left(\left(y^{\prime}_2 - y_1\right)x^{\prime}_2 - \left(x^{\prime}_2 - x_1\right)y^{\prime}_2\right)b_z = 0
\end{split}
\end{equation}
We therefore obtain a linear equation in the components of $b$ which can be used to formulate an overdetermined linear system that can be solved in ordinary least squares fashion. It should be stressed here that eq. \ref{eq:LS_equation_without_depth} is tolerant to points with very low disparity as it is quite evident that, if both terms $ \left(x^{\prime}_2 - x_1\right)$  and $\left(y^{\prime}_2 - y_1\right)$ vanish, then the equation becomes trivial $\left(0b=0\right)$ and subsequently has little or no effect on the least squares estimate. We may now formally "repackage" eq. \ref{eq:LS_equation_without_depth} for the i\textsuperscript{th} feature in a new camera view at time $t$ using a determinant:
\begin{equation}
\label{eq:determinant_LS_equation}
\det\begin{bmatrix}R_{h_i}\left(s_t - s_{h_i}\right) & \left(m^{\left(i\right)}_t\right)^{\prime}-m^{\left(i\right)}_{h_i} & \left(m^{\left(i\right)}_t\right)^{\prime}\end{bmatrix} = 0
\end{equation} 
where $h_i$ is the time index of the camera view in which the i\textsuperscript{th} feature was originally detected,  $s_t$ and $s_{h_i }$ are the position of the camera at times $t$ and $h_i$ in world coordinates, $R_{h_i}$ is the orientation matrix of the camera at time $h_i$, $m^{\left(i\right)}_{h_i}$ is the normalized Euclidean projection of the feature in its home camera view and ${\left(m^{(i)}_t\right)}^{'}$  is the “unrotated” normalized Euclidean projection in the camera view captured at time t.
\par In the initializing pair of views (i.e., for $h_i=0$ and $t=1$), the overdetermined linear system will obviously be homogeneous and therefore the solution will be the direction of the baseline between the respective camera frames. In this case, of multiple camera views, the solution will be sign-ambiguous and therefore it will be necessary to reconstruct the scene and vote for the best reconstruction (i.e., the one with fewer negative/vanishing depths). However, in the case of more than two camera views, the system becomes a standard least squares formulation with a unique scale-aware solution in which scale is infused by the known positions of previous camera frames.

\subsection{Outlier rejection}
To eliminate outliers we employ a robust random sample consensus \cite{fischler1981random} scheme called MLESAC proposed by Torr and Zisserman \cite{torr2000mlesac}. For each new camera view, we compute the baseline using eq. \ref{eq:determinant_LS_equation} over randomly picked minimal subsets of at least 3 points. Then, for the recovered camera position we obtain a robust epipolar score on the misalignment of epipolar planes between camera views. Unlike the case of Sampson error \cite{hartley2003multiple}, our epipolar misalignment error works directly on the relative pose without the need to use an essential matrix and it is generally easy to compute.
\begin{figure}[ht]
\centering
\centerline{\includegraphics[width=\textwidth,keepaspectratio]{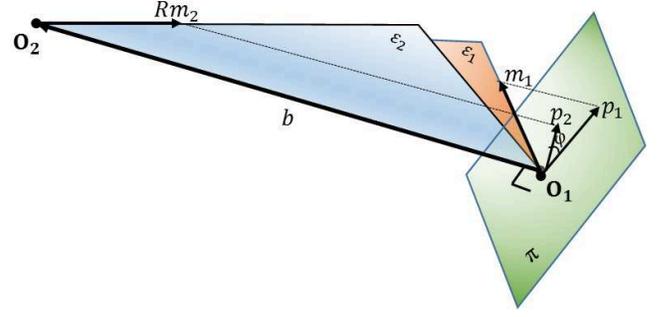}}
\caption{Measuring misalignment of epipolar planes $\epsilon_1$ and $\epsilon_2$ on a plane $\pi$, orthogonal to the baseline $b$.The misalignment angle cosine will be the inner-product of the normalized projections of $m_1$ and $Rm_2$ on the plane $\pi$. }
\label{fig:epipolar_plane_misalignement}       % Give a unique label
\end{figure}
\par Consider the two corresponding normalized Euclidean projections $m_1$ and $m_2$ (again, superscripts are dropped for simplicity) in two camera views indexed by 1 and 2. Then, if the correspondences are noisy, the two epipolar planes defined by $m_1$, $m_2$ and the baseline will not coincide as shown in Figure \ref{fig:epipolar_plane_misalignement} and will subsequently form an angle between them. Since the baseline vector $b$ is common for both planes, one way of obtaining the angle between the two misaligned planes is to project the two correspondences onto the orthogonal space of the baseline, which is essentially a plane $\pi$ through the origin $O_1$ and to which normal is the baseline. The (projector) matrix $P$ that projects a vector onto the space orthogonal to $b$ is:
\begin{equation}
\label{eq:baseline_orthogonal_projector}
P = I_3 - bb^T
\end{equation}
Thus, the projection of $m_1$ would be:
\begin{equation}
\label{eq:m1_orthogonal_projection}
p_1 = Pm_1 = (I_3 - bb^T)m_1
\end{equation}

The projection of $m_2$, accounting for the difference in the orientation of camera frames, is:

\begin{equation}
\label{eq:m2_orthogonal_projection}
p_2 = PRm_2 = (I_3 - bb^T)Rm_2
\end{equation}
where $R$ is the relative orientation matrix. The cosine of the angle between the two planes $\epsilon_1 = <m_1,b>$ and $\epsilon_2 = <Rm_2,b>$ will be the following inner-product:

\begin{equation}
\label{eq:cosine_between_epipolar_planes}
\begin{aligned}
\cos{\phi} &= \frac{p^T_1 p_2}{\Vert p_1\Vert \Vert p_2\Vert} \\
&= \frac{m^T_2R^T\left(I_3 - bb^T\right)m_1}{\sqrt{{\Vert m_1\Vert}^2-{\left(m^T_1b\right)}^2}\sqrt{\Vert m_2\Vert^2-{\left(m^T_2R^Tb\right)}^2}}
\end{aligned}
\end{equation}
\par Clearly, the measure of eq. \ref{eq:cosine_between_epipolar_planes} is a score that increases with the accuracy of the correspondence and therefore we seek to maximize it. Nominal values used for the MLESAC cutoff bound were in the range of $\cos{7^\circ}$ to $\cos{3^\circ}$. It should be stressed here that the score of eq. \ref{eq:cosine_between_epipolar_planes} penalizes angles above 90\textsuperscript{$\circ$} (i.e., correspondences with negative depth in exactly one view), whereas the classic epipolar constraint does not (due to the minimization requirement). Most importantly, it is a score that is independent of the type of camera motion and fitted model (homography or essential matrix), including the case of pure rotational camera motion, in which case the two projection directions should be collinear and the formula reduces to a simple inner-product which yields the cosine of the angle between $m_1$ and $Rm_2$.
\subsection{Iterative refinement}
Pose estimates are iteratively refined with the Gauss-Newton method over a few (usually 3) recent views. Instead of the traditional reprojection error \cite{hartley2003multiple}, we employ an epipolar alignment score, similar to the one of eq. \ref{eq:cosine_between_epipolar_planes}. In particular, instead of considering the misalignment of epipolar planes, this time, we focus the angle between the direction of the second correspondence and the normal of the first plane. This way, a minimization problem is obtained, suitable to apply the Gauss-Newton method. Suppose a sliding window of $n$ camera views is used. We construct the following cost function $C$ over this window:
\begin{equation}
\label{eq:epipolar_reprojection_error}
C=\sum^{t+n-1}_{k = t}{\sum_{i\in V_t}{\left(m_tR^T_t\left[s_t-s_{h_i}\right]_{\times}R_{h_i}m_{h_i}\right)^2}}
\end{equation}
where $V_t$ is the set of visible points in the window commencing at time $t$, $h_i$ is the time index of the "home" view (i.e., first detection) of the i\textsuperscript{th} feature, $m^{\left(i\right)}_{t}$ is the normalized Euclidean projection of the i\textsuperscript{th} feature in the camera view at time $t$ and $R_k$, $s_k$ are the orientation matrix and position vector (in world coordinates) of the camera at time $t=k$.
\par The cost function of eq. \ref{eq:epipolar_reprojection_error} is a sum of epipolar constraints over the camera poses in the sliding window. The latter suggests that the Gauss-Newton normal equations scale only with camera poses, as opposed to the larger system sizes we typically obtain with sparse bundle adjustment. For a window of size $n=2$, the cost function is {\it scale unaware} and becomes the{\it essential matrix} least squares formulation \cite{terzakis2012relativepose}. For $n>2$, at least one camera pose is taken as constant and the optimization problem becomes {\it scale aware} owed to the presence of measurements in more than two camera views.

\section{Experiments}
We executed our method on video sequences of the river shore recorded on a cruising vehicle averaging 5-7 knots speed. Each sequence is accompanied by IMU samples at 150 Hz and GPS readings at approximately 1Hz (standard refresh rate of USB dongle). Both IMU and GPS data are time-stamped with a video frame index. The GPS readings are used as our main measure of "ground truth". 

\subsection{Ground truth}
\label{sect:interpolation_for_ground_truth}
Due to the limitations imposed primarily by the environment, GPS was the only reliable means of a ground truth estimate. Although the GPS position reading is known to have approximately 1.5 meters variance, it however is highly reliable in long routes, simply because it does not accumulate error. The routes corresponding to the recorded sequences are longer than 30 meters and therefore, in this context, the GPS routes are accurate representations of ground truth.
\par The scale of the vision-based estimated trajectory is defined by the last reading of speed over ground from the GPS, before the visual odometer takes over. Since this is a fairly noisy estimate and scale errors accumulate during SLAM, direct comparisons between the GPS trajectory and the respective SLAM-generated estimate would be biased, regardless of whether this estimate is adequate to safely localize the vehicle while cruising. Thus, in order to perform unbiased comparisons, both trajectories are interpolated with a Catmull – Rom spline \cite{catmull1974class} and parametrized relative to their length; using the common parameter, we estimate the optimal (in the least squares sense) affine transformation that minimizes the distances between the two splines for common parameter values. The choice of interpolating spline can be arbitrary, but Catmull-Rom curves are more convenient to construct locally with only four points at a time. It should be stressed that in order to obtain corresponding parameters values between the GPS and vision based trajectories, it is necessary to re-parametrize the curves by means of arc length. The general concept is illustrated in Figure 8.
\textbf{\begin{figure}[ht]
\centering
\centerline{\includegraphics[width=\textwidth,height=.6\textwidth]{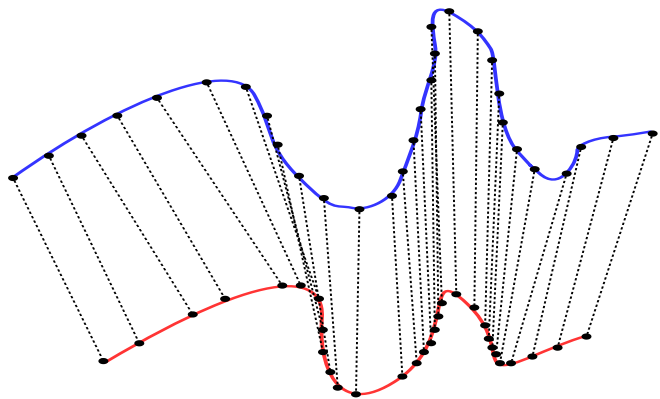}}
\caption{Parameter correspondence between the GPS (blue) and visual odometry (red) interpolated curves irrespectively of scale. Synchronization is achieved using the GPS-frame index log.}
\label{fig:interpolation_parameter_correspondence}       % Give a unique label
\end{figure}
}
\par Suppose we fit two splines $g\left(t_g\right)$ and $v\left(t_v\right)$ where $t_g, t_v \in \left[0, 1\right]$ to both GPS and vision based odometry points using a common parameter $u\in\left[0, 1\right]$. It is necessary for this parameter to reflect percentage of overall arc-length in either curve. Thus, the warping functions for $t_g \left(u\right)$ and $t_v \left(u\right)$ can be computed from the following differential equations:

\begin{eqnarray}
\label{eq:arclength_g}
s_g\left(u\right) &=& \int_{0}^{t_g \left(u\right)} {g\left(t\right)\,dt}\\
\label{eq:arclength_v}
s_v\left(u\right) &=& \int_{0}^{t_v \left(u\right)} {v\left(t\right)\,dt}
\end{eqnarray}
where $s_g$ and $s_v$ are the arc-lengths of $g$ and $h$ respectively. Having obtained the reparametrized splines in terms of $u$, it is now a matter of ordinary least squares to fit a 2D affine transformation $A$ that minimizes the distance between a number of sampled points in $g$ and $v$ for common values of $u$:
\begin{equation}
\label{eq19}
\hat{A} = \underset{A}{\mathrm{argmin}}\Bigg\{ \sum_{k=0}^{n-1} {\Bigg\Vert A\,v\left(t_v\left(\frac{k}{n-1}\right)\right) - g\left(t_g\left(\frac{k}{n-1}\right)\right)\Bigg\Vert} ^2 \Bigg\}
\end{equation}
where $n$ is the number of uniform samples in [0,1].

\subsection{Odometry estimates}
\begin{figure}[ht]
\centering
\centerline{\includegraphics[width=\textwidth,height=.6\textwidth]{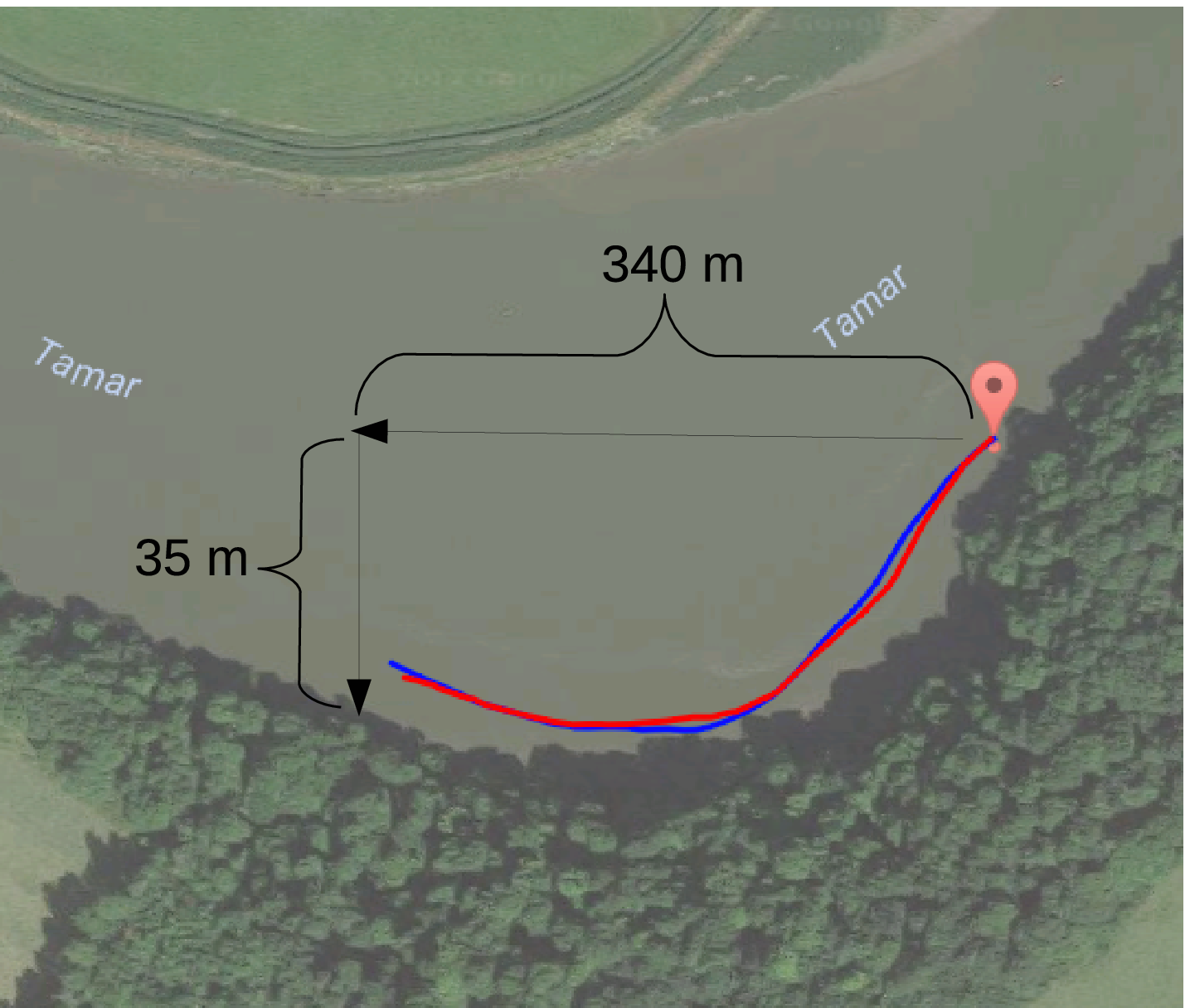}}
\caption{GPS ground truth spline (blue) and estimated visual odometry spline (red) overlaid in satellite imagery. Approximate distances along the $x$ and $y$ axes are given to indicate scale.}
\label{fig:overlay_21_05_2013_10_21_splines}       % Give a unique label
\end{figure}

\begin{figure}[ht]
\centering
\centerline{\includegraphics[width=\textwidth,height=.6\textwidth]{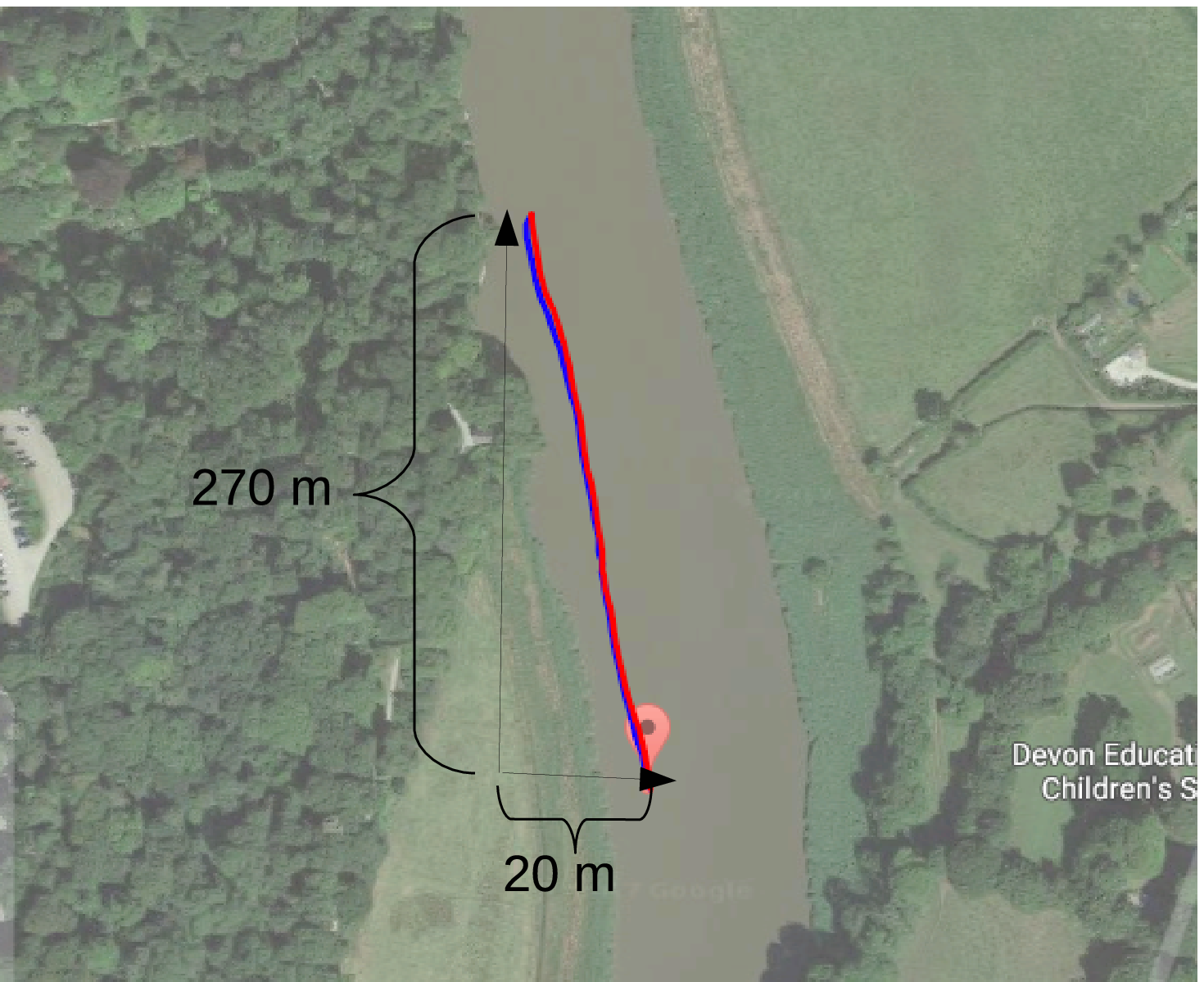}}
\caption{GPS ground truth spline (blue) and estimated visual odometry spline (red) overlaid in satellite imagery. Approximate distances along the $x$ and $y$ axes are given to indicate scale.}
\label{fig:overlay_21_05_2013_10_35_splines}       % Give a unique label
\end{figure}

Figures \ref{fig:overlay_21_05_2013_10_21_splines} and \ref{fig:overlay_21_05_2013_10_35_splines} illustrate odometry estimates and GPS ground truth superimposed on satellite imagery for two indicative sequences. Overall data was recorded in the Tamar river, Devon, UK. The trajectories are generally long (several hundred meters); unfortunately, the vehicle had to follow specific routes in order to avoid shallow waters and this many times resulted in trajectories without many twists. It should be noted however that transient rotational and linear motion is generally rich as shown in the actual footage, yet it is not reflected in the GPS based trajectory for obvious reasons. 
%\par Although the method is structure-less, a reconstruction can be obtained from the measured projections and the camera pose estimates. The resulting map shown in Figures 9-13 is purely for visualization purposes as it does not play a role in camera pose estimation. Many distant features are included in the map, as they were not classified as outliers by the RANSAC algorithm described in section 3.2; this way, we may actually observe the detrimental effect that some of these map-points could have on the camera resectioning process due to accumulated uncertainty in the position estimates.  <I think the figures need to be redone. I don’t like them very much. Also, some map-points are really bad, and maybe we can throw them away manually in order to demonstrate the strong side of the 3D reconstruction >
\par
\begin{figure}
\centering
  \includegraphics[width=1.\linewidth,height=.7\linewidth]{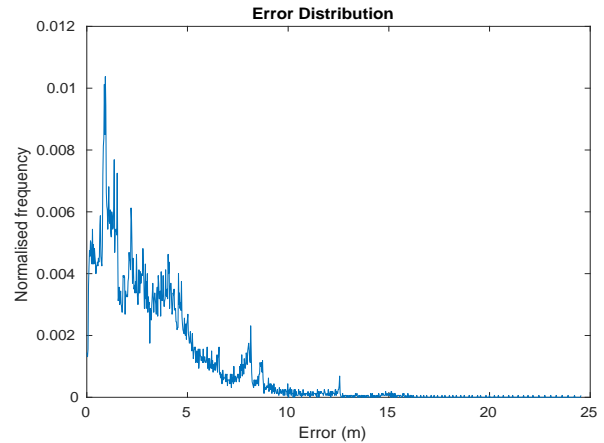}
\caption{Position error distribution obtained from 8 routes in the river.}

\label{fig:overall_error_distribution}
\end{figure}

 We evaluated the quality of localization in a total of 8 trajectories, averaging approximately 350m total distance covered on the water. The position error was evaluated in meters using the method described in Section \ref{sect:interpolation_for_ground_truth}. On average, the maximum position error is typically in the range of 8-10m with the exception of occasional occurrences in the 20m range, such as in the 400m long trajectory of Figure \ref{fig:overlay_13_01_2014_11_10}. The position error distribution obtained from all 8 sequences is depicted in Figure \ref{fig:overall_error_distribution}. Interestingly, the plot is very reminiscent of the $\chi^2$ distribution, which empirically indicates consistency with normally distributed relative pose squared error.

\begin{figure*}
\centering
\begin{subfigure}{.33\textwidth}
  \centering
  \includegraphics[width=1.\linewidth]{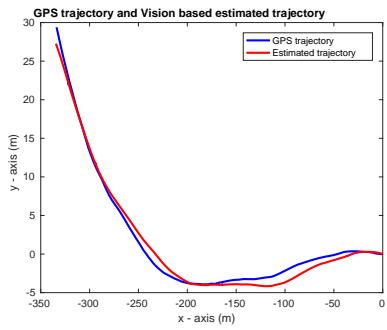}
  \caption{Visual odometry vs GPS trajectory.}
  \label{fig:sub11}
\end{subfigure}%
\begin{subfigure}{.33\textwidth}
  \centering
  \includegraphics[width=1.\linewidth]{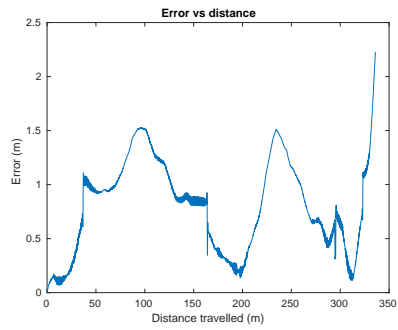}
  \caption{Position error vs Distance travelled.}
  \label{fig:sub12}
\end{subfigure}
\begin{subfigure}{.33\textwidth}
  \centering
  \includegraphics[width=1.\linewidth]{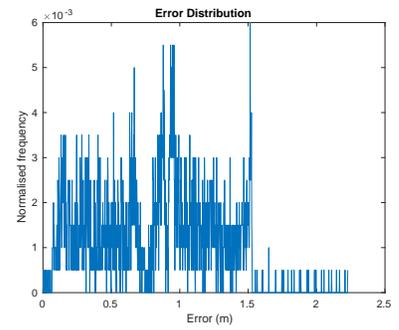}
  \caption{Error distribution.}
  \label{fig:sub13}
\end{subfigure}

\caption{An approximately 340m long route along the Tamar river, near Halton Quay, Devon, UK.}

\label{fig:overlay_21_05_2013_10_21}
\end{figure*}

\begin{figure*}
\centering
\begin{subfigure}{.33\textwidth}
  \centering
  \includegraphics[width=1.\linewidth]{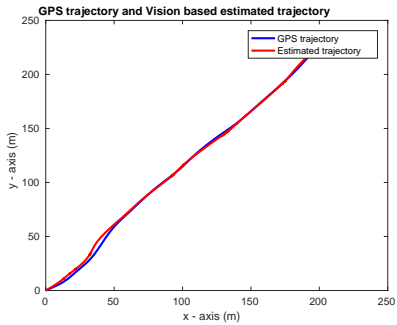}
  \caption{Visual odometry vs GPS trajectory.}
  \label{fig:sub21}
\end{subfigure}%
\begin{subfigure}{.33\textwidth}
  \centering
  \includegraphics[width=1.\linewidth]{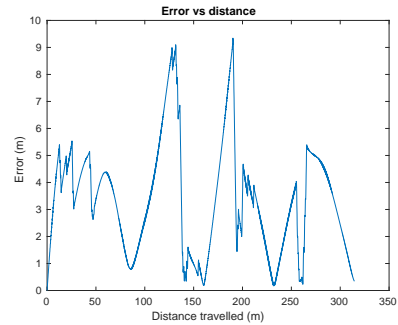}
  \caption{Position error vs Distance travelled.}
  \label{fig:sub22}
\end{subfigure}
\begin{subfigure}{.33\textwidth}
  \centering
  \includegraphics[width=1.\linewidth]{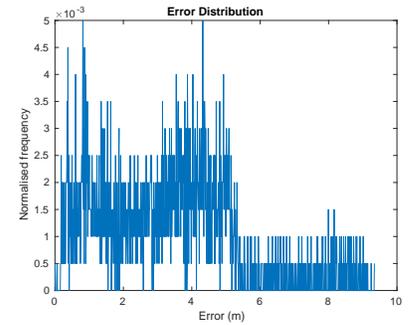}
  \caption{Error distribution.}
  \label{fig:sub23}
\end{subfigure}

\caption{An approximately 285m long route along the Tamar river, in Bohetherick, Devon, UK.}

\label{fig:overlay_21_05_2013_10_35}
\end{figure*}

\begin{figure*}
\centering
\begin{subfigure}{.33\textwidth}
  \centering
  \includegraphics[width=1.\linewidth]{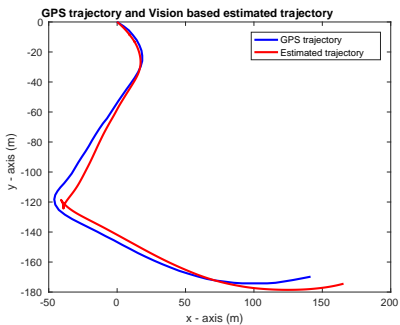}
  \caption{Visual odometry vs GPS trajectory.}
  \label{fig:sub31}
\end{subfigure}%
\begin{subfigure}{.33\textwidth}
  \centering
  \includegraphics[width=1.\linewidth]{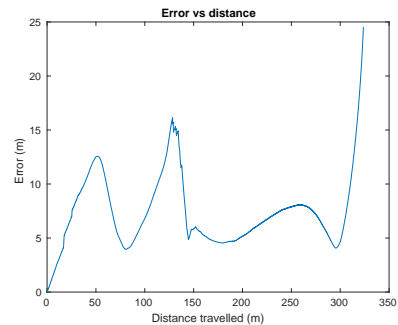}
  \caption{Position error vs Distance travelled.}
  \label{fig:sub32}
\end{subfigure}
\begin{subfigure}{.33\textwidth}
  \centering
  \includegraphics[width=1.\linewidth]{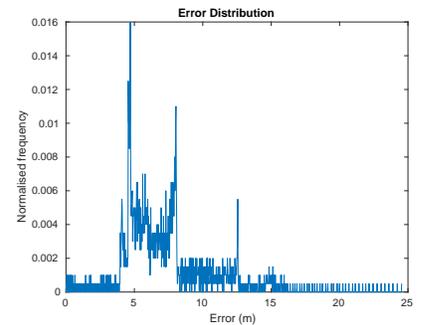}
  \caption{Error distribution.}
  \label{fig:sub33}
\end{subfigure}

\caption{An approximately 400m long route along the Tamar river, in Calstock, Devon, UK.}

\label{fig:overlay_13_01_2014_11_14}
\end{figure*}

\begin{figure*}
	\centering
	\begin{subfigure}{.33\textwidth}
		\centering
		\includegraphics[width=1.\linewidth]{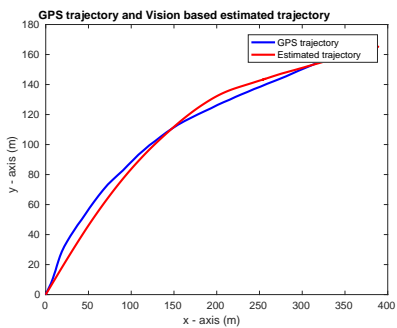}
		\caption{Visual odometry vs GPS trajectory.}
		\label{fig:sub41}
	\end{subfigure}%
	\begin{subfigure}{.33\textwidth}
		\centering
		\includegraphics[width=1.\linewidth]{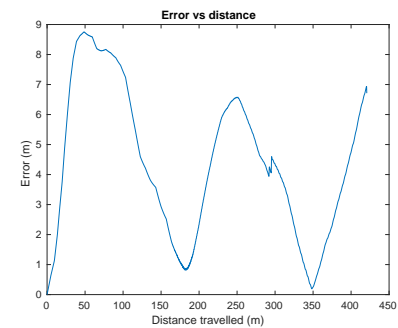}
		\caption{Position error vs Distance travelled.}
		\label{fig:sub42}
	\end{subfigure}
	\begin{subfigure}{.33\textwidth}
		\centering
		\includegraphics[width=1.\linewidth]{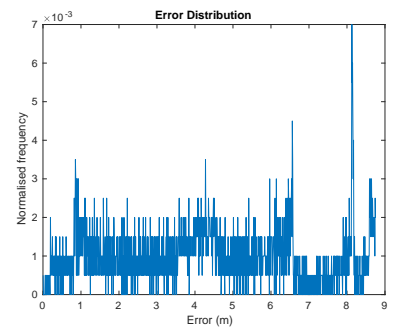}
		\caption{Error distribution.}
		\label{fig:sub43}
	\end{subfigure}
	
	\caption{An approx6imately 450m long route along the Tamar river, near Cotehele, Devon, UK.}
	
	\label{fig:overlay_13_01_2014_11_10}
\end{figure*}

Figures \ref{fig:overlay_21_05_2013_10_21}, \ref{fig:overlay_21_05_2013_10_35}, \ref{fig:overlay_13_01_2014_11_10}, \ref{fig:overlay_13_01_2014_11_14} illustrate plots generated from 4 selected sequences, illustrating comparison of visual odometry with ground truth, position error distribution and its progression with distance traveled by the vehicle.  

\section{Conclusion}
We have presented a method to obtain reliable visual odometry estimates for an unmanned sea-surface vehicle using imagery of the shore, in order to accommodate autonomous cruise when GPS feedback becomes unavailable. To the best of our knowledge, this is the first time that a sea-surface vehicle location is obtained exclusively with standard Visual SLAM techniques. Except for the typical shortcomings of outdoor visual SLAM (brightness, shading, motion blur, environmental conditions, etc.), the application at hand presents us with additional challenges associated with the absence of a ground plane and extreme scene-depth variation. Typically, the ground plane endows the imagery with features that correspond to near-by world points based on which, camera motion can be reliably estimated. In contrast, for imagery obtained on a cruising boat, there is no ground plane and motion must be estimated from features in the background which may or may not be near, as depth varies significantly.

\par To cope with depth variation in the useful portion of the background, we advocate the use of structure-less visual SLAM with the aid of orientation priors from an IMU. With this information, we are able to cast the relative pose problem directly in terms of image projections, thereby circumventing the map which may contain a number of noisy reconstructions. Our relative pose equations do not discard distant points, but they simple ignore them when their disparity is not large enough to provide additional motion information. However, regardless of depth, a point can always be an outlier, so in order to combine our model with RANSAC, we propose a robust error reflecting epipoplar plane misalignment. The error is valid for any type of degenerate configuration, including pure rotational motion. To avoid having to use the map during error (reprojection) refinement, we employ a cost function which is formed by the sum of standard epipolar constraints.

\par The specifics of the application described herein, naturally do not exactly match the conditions in available datasets. We therefore recorded our own sequences and timestamped inertial data and GPS readings with frame indexes in each sequence. To obtain unbiased comparisons with ground truth, it was necessary to compare the GPS-generated trajectory with visual odometry using a common parameter that reflects equal proportions of arc-length with respect to the origin in both curves. To do so, we interpolate both curves using a spline and thereafter, reparametrize resulting curves in terms of their arc-length in the interval $\left[0,1\right]$. The use of a common arc-length parameter allows for actual-scale comparisons with the GPS trajectory.


%%%%%%%%%%%%%%%%%%%%%%%%%%%%%%%%%%%%%%%%%%%%%%%%%%%%%%%%%%%%%%%%%%%%%%%%%%%%%%%%

%%%%%%%%%%%%%%%%%%%%%%%%%%%%%%%%%%%%%%%%%%%%%%%%%%%%%%%%%%%%%%%%%%%%%%%%%%%%%%%%

%%%%%%%%%%%%%%%%%%%%%%%%%%%%%%%%%%%%%%%%%%%%%%%%%%%%%%%%%%%%%%%%%%%%%%%%%%%%%%%%

%%%%%%%%%%%%%%%%%%%%%%%%%%%%%%%%%%%%%%%%%%%%%%%%%%%%%%%%%%%%%%%%%%%%%%%%%%%%%%%%

\bibliographystyle{IEEEtran}      % basic style, author-year citations
\bibliography{IEEEexample}   % name your BibTeX data base

\end{document}